\documentclass{article}

\usepackage{times}
\usepackage{graphicx} 
\usepackage{subfigure}

\usepackage{natbib}

\usepackage{algorithm}
\usepackage{algorithmic}

\usepackage{color,mathtools}
\usepackage{amssymb,amsfonts}
\usepackage{amsmath,amsthm}
\usepackage{float}
\usepackage{epsfig}
\usepackage{comment}
\usepackage{verbatim}
\usepackage{appendix}
\usepackage[all,dvips,arc,curve,color,frame]{xy}
\usepackage{bbm}
\usepackage{wrapfig}
\usepackage{dsfont}
\usepackage[T1]{fontenc}
\usepackage{xargs}
\usepackage[usenames,dvipsnames]{xcolor}

\newcommand{\argmin}{\operatornamewithlimits{argmin}}

\usepackage[colorlinks=true, allcolors=blue]{hyperref}

\usepackage[accepted]{icml2017}

\icmltitlerunning{Adaptive Neural Networks for Efficient Inference}

\begin{document}

\twocolumn[
\icmltitle{Adaptive Neural Networks for Efficient Inference}



\icmlsetsymbol{equal}{*}

\begin{icmlauthorlist}
\icmlauthor{Tolga Bolukbasi}{bu}
\icmlauthor{Joseph Wang}{am}
\icmlauthor{Ofer Dekel}{msr}
\icmlauthor{Venkatesh Saligrama}{bu}
\end{icmlauthorlist}

\icmlaffiliation{bu}{Boston University, Boston, MA, USA}
\icmlaffiliation{am}{Amazon, Cambridge, MA, USA}
\icmlaffiliation{msr}{Microsoft Research, Redmond, WA, USA}

\icmlcorrespondingauthor{Tolga Bolukbasi}{tolgab@bu.edu}

\icmlkeywords{deep neural networks, object recognition, budgeted learning, resource efficient prediction, conditional computation, efficient inference}

\vskip 0.3in
]



\printAffiliationsAndNotice{}  

\begin{abstract} 
We present an approach to adaptively utilize deep neural networks in order to reduce the evaluation time on new examples without loss of accuracy. Rather than attempting to redesign or approximate existing networks, we propose two schemes that adaptively utilize networks. We first pose an adaptive network evaluation scheme, where we learn a system to adaptively choose the components of a deep network to be evaluated for each example. By allowing examples correctly classified using early layers of the system to exit, we avoid the computational time associated with full evaluation of the network. We extend this to learn a network selection system that adaptively selects the network to be evaluated for each example. We show that computational time can be dramatically reduced by exploiting the fact that many examples can be correctly classified using relatively efficient networks and that complex, computationally costly networks are only necessary for a small fraction of examples.
We pose a global objective for learning an adaptive early exit or network selection policy and solve it by reducing the policy learning problem to a layer-by-layer weighted binary classification problem.  Empirically, these approaches yield dramatic reductions in computational cost, with up to a 2.8x speedup on state-of-the-art networks from the ImageNet image recognition challenge with minimal ($<1\%$) loss of top5 accuracy.
\end{abstract}

\section{Introduction}
\begin{figure}[!b]
	\vspace{-10pt}
   \begin{center}
    \includegraphics[width=\linewidth]{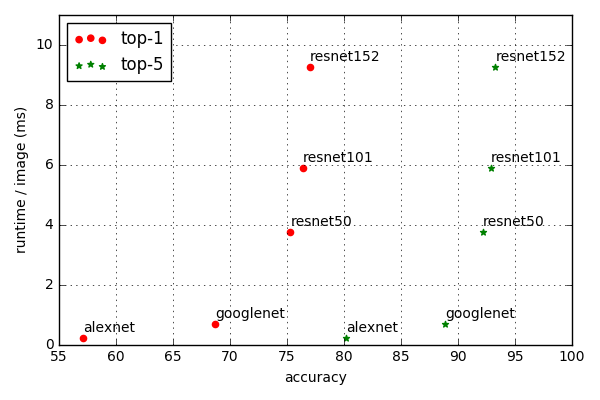}
  \end{center}
  \vspace{-12pt}
  \caption[Network tradeoff]{\small Performance versus evaluation complexity of the DNN architectures that won the ImageNet challenge over past several years. The model evaluation times increase exponentially with respect to the increase in accuracy.}
  \label{fig:imagenet_challenge}
\end{figure}

Deep neural networks (DNNs) are among the most powerful and versatile machine learning techniques, achieving state-of-the-art accuracy in a variety of important applications, such as visual object recognition \cite{he2016deep}, speech recognition \cite{hinton2012deep}, and machine translation \cite{sutskever2014sequence}. However, the power of DNNs comes at a considerable cost, namely, the computational cost of applying them to new examples. This cost, often called the \emph{test-time cost}, has increased rapidly for many tasks (see Fig.~\ref{fig:imagenet_challenge}) with ever-growing demands for improved performance in state-of-the-art systems. As a point of fact, the \textit{Resnet152} \cite{he2016deep} architecture with 152 layers, realizes a substantial 4.4\% accuracy gain in top-5 performance over GoogLeNet \cite{szegedy2015going} on the large-scale ImageNet dataset \cite{russakovsky2015imagenet} but is about 14X slower at test-time.  
The high test-time cost of state-of-the-art DNNs means that they can only be deployed on powerful computers, equipped with massive GPU accelerators. As a result, technology companies spend billions of dollars a year on expensive and power-hungry computer hardware. Moreover, high test-time cost prevents DNNs from being deployed on resource constrained platforms, such as those found in Internet of Things (IoT) devices, smart phones, and wearables.
This problem has given rise to a concentrated research effort to reduce the test-time cost of DNNs. Most of the work on this topic focuses on designing more efficient network topologies and on compressing pre-trained models using various techniques (see related work below). We propose a different approach, which leaves the original DNN intact and instead changes the way in which we apply the DNN to new examples. We exploit the fact that natural data is typically a mix of easy examples and difficult examples, and we posit that the easy examples do not require the full power and complexity of a massive DNN.

We pursue two concrete variants of this idea. First, we propose an adaptive early-exit strategy that allows easy examples to bypass some of the network's layers. Before each expensive neural network layer (e.g., convolutional layers), we train a policy that determines whether the current example should proceed to the next layer, or be diverted to a simple classifier for immediate classification. Our second approach, an adaptive network selection method, takes a set of pre-trained DNNs, each with a different cost/accuracy trade-off, and arranges them in a directed acyclic graph \cite{trapeznikov:2013b,wang2015efficient}, with the the cheapest model first and the most expensive one last. We then train an exit policy at each node in the graph, which determines whether we should rely on the current model's predictions or predict the most beneficial next branch to forward the example to. In this context we pose a global objective for learning an adaptive early exit or network selection policy and solve it by reducing the policy learning problem to a layer-by-layer weighted binary classification problem.

We demonstrate the merits of our techniques on the ImageNet object recognition task, using a number of popular pretrained DNNs. The early exit technique speeds up the average test-time evaluation of GoogLeNet \cite{szegedy2015going}, and Resnet50 \cite{he2016deep} by 20-30\% within reasonable accuracy margins. The network cascade achieves 2.8x speed-up compared to pure Resnet50 model at 1\% top-5 accuracy loss and 1.9x speed-up with no change in model accuracy. We also show that our method can approximate a oracle policy that can see true errors suffered for each instance. 

In addition to reducing the average test-time cost of DNNs, it is worth noting that our techniques are compatible with the common design of large systems of mobile devices, such as smart phone networks or smart surveillance-camera networks. These systems typically include a large number of resource-constrained edge devices that are connected to a central and resource-rich cloud. One of the main challenges involved in designing these systems is determining whether the machine-learned models will run in the devices or in the cloud. Offloading all of the work to the cloud can be problematic due to network latency, limited cloud ingress bandwidth, cloud availability and reliability issues, and privacy concerns. Our approach can be used to design such a system, by deploying a small inaccurate model and an exit policy on each device and a large accurate model in the cloud. Easy examples would be handled by the devices, while difficult ones would be forwarded to the cloud. Our approach naturally generalizes to a fog computing topology (where resource constrained edge devices are connected to a more powerful local gateway computer, which in turn is connected to a sequence of increasingly powerful computers along the path to the data-center). Such designs allow our method to be used in memory constrained settings as well due to offloading of complex models from the device.

\begin{figure*}[!t]
	\vspace{-5pt}
	\begin{center}
    \includegraphics[width=0.32\linewidth]{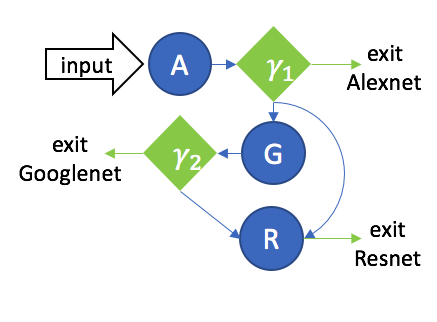}            		\includegraphics[width=0.67\linewidth]{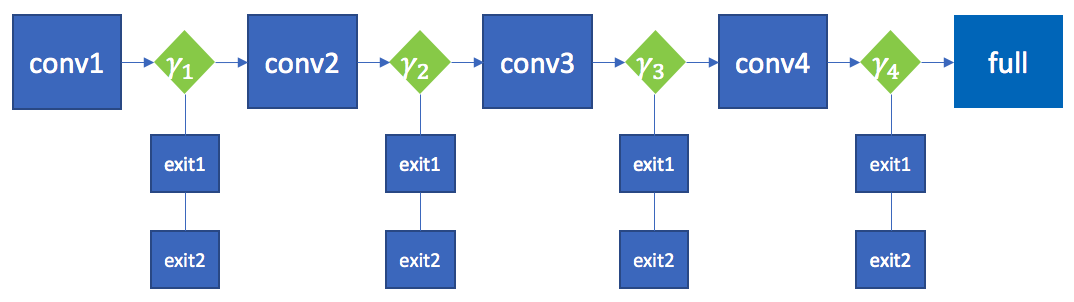}
  \end{center}
  \vspace{-13pt}
  \caption[Network selection architecture]{\textbf{(Left)} An example network selection system topology for networks Alexnet(A), GoogLeNet(G) and Resnet(R). Green $\gamma$ blocks denote the selection policy. The policy evaluates Alexnet, receives confidence feedback and decides to jump directly to Resnet or send the sample to GoogLeNet->Resnet cascade. \textbf{(Right)} An example early exit system topology (based on Alexnet). The policy chooses one of the multiple exits available to it at each stage for feedback. If the sample is easy enough, the system sends it down to exit, otherwise it sends the sample to the next layer.}
  \label{fig:network_selection_arch}
  \vspace{-5pt}
\end{figure*}

\section{Related Work}
Past work on reducing evaluation time of deep neural networks has centered on reductions in precision and arithmetic computational cost, design of efficient network structure, and compression or sparsification of networks to reduce the number of convolutions, neurons, and edges. The approach proposed in this paper is complimentary. Our approach does not modify network structure or training and can be applied in tandem with these approaches to further reduce computational cost.

The early efforts to compress large DNNs used a large \emph{teacher model} to generate an endless stream of labeled examples for a smaller \emph{student model} \cite{bucilua2006model,hinton2015distilling}.
The wealth of labeled training data generated by the teacher model allowed the small student model to mimic its accuracy.

Reduced precision networks \cite{gong2014compressing,courbariaux2015binaryconnect,chen2015compressing,hubara2016binarized,wu2016quantized,rastegari2016xnor,hubara2016quantized}
have been extensively studied to reduce the memory footprint of networks and their test-time cost.
Similarly, computationally efficient network structures have also been proposed to reduce the computational cost of deep networks by exploiting efficient operations to approximate complex functions, such as the inception layers introduced in GoogLeNet \cite{szegedy2015going}.

Network sparsification techniques attempt to identify and prune away redundant parts of a large neural networks. A common approach is to remove unnecessary nodes/edges from the network\cite{liu2015sparse,iandola2016squeezenet,wen2016learning}. In convolutional neural networks, the expensive convolution layers can be approximated \cite{bagherinezhad2016lcnn} and redundant computation can be avoided \cite{figurnov2016perforatedcnns}.

More recently, researchers have designed spatially adaptive networks \cite{figurnov2016spatially, bengio2015conditional} where nodes in a layer are selectively activated. Others have developed  cascade approaches \cite{leroux2017cascading,odena2017changing} that allow early exits based on confidence feedback. Our approach can be seen as an instance of conditional computation, where we seek computational gains through layer-by-layer and network-level early exits. However, we propose a general framework which optimizes a novel system risk that includes computational costs as well as accuracy. Our method does not require within layer modifications and works with directed acyclic graphs that allow multiple model evaluation paths.

Our techniques for adaptive DNNs borrow ideas from the related sensor selection problem \cite{xu2013cost,kusner2014feature,wang2014model,wang2015efficient,trapeznikov:2013b,NanWS16,wang2012local}. The goal of sensor selection is to adaptively choose sensor measurements or features for each example.

\section{Adaptive Early Exit Networks}
Our first approach to reducing the test-time cost of deep neural networks is an early exit strategy. We first frame a global objective function and reduce policy training for optimizing the system-wide risk to a layer-by-layer {\it weighted binary classification (WBC)}. We denote a labeled example as $(x,y) \in \mathds{R}^{d}\times \{1,\ldots,\mathcal{L}\}$, where $d$ is the dimension of the data and $\{1,\ldots,\mathcal{L}\}$ is the set of classes represented in the data. We define the distribution generating the examples as $\mathcal{X}\times\mathcal{Y}$. For a predicted label $\hat{y}$, we denote the loss $L(\hat{y},y)$. In this paper, we focus on the task of classification and, for exposition, focus on the indicator loss $L(\hat{y},y)=\mathds{1}_{\hat{y}=y}$, in this section. In practice we upper bound the indicator functions with logistic loss for computational efficiency. %

As a running DNN example, we consider the AlexNet architecture \cite{krizhevsky2012imagenet}, which is composed of 5 convolutional layers followed 3 fully connected layers. During evaluation of the network, computing each convolutional layer takes more than 3 times longer than computing a fully connected layer, so we consider a system that allows an example to exit the network after each of the first 4 convolutional layers. Let $\hat{y}(x)$ denote the label predicted by the network for example $x$ and assume that computing this prediction takes a constant time of $T$. Moreover, let $\sigma_k(x)$ denote the output of the $k^{\text{th}}$ convolutional layer for example $x$ and let $t_k$ denote the time it takes to compute this value (from the time that $x$ is fed to the input layer).
Finally, let $\hat{y}_{k}(x)$ be the predicted label if we exit after the $k^{\text{th}}$ layer.

After computing the $k^{\text{th}}$ convolutional layer, we introduce a decision function $\gamma_k$ that determines whether the example should exit the network with a label of $\hat{y}_{k}(x)$ or proceed to the next layer for further evaluation. The input to this decision function is the output of the corresponding convolutional layer $\sigma_k(x)$, and the value of $\gamma_k(\sigma_k(x))$ is either $-1$ (indicating an early exit) or $1$. This architecture is depicted on the right-hand side of Fig. \ref{fig:network_selection_arch}.

Globally, our goal is to minimize the evaluation time of the network such that the error rate of the adaptive system is no more than some user-chosen value $B$ greater than the full network:
\begin{align}
&\min_{\gamma_1,...,\gamma_4} \mathds{E}_{x\sim \mathcal{X}}\left[T_{\gamma_1,\ldots,\gamma_4}(x)\right].\\
&\mbox{s.t.  } \mathds{E}_{(x,y)\sim\mathcal{X}\times\mathcal{Y}}\left[\left(L(\hat{y}{\gamma_1,...,\gamma_4}(x),y)-L(\hat{y}(x),y)\right)_+\right]\leq B\nonumber
\end{align}
Here, $T_{\gamma_1,...,\gamma_4}(x)$ is the prediction time for example $x$ for the adaptive system, $\hat{y}{\gamma_1,...,\gamma_4}(x)$ is the label predicted by the adaptive system for example $x$. In practice, the time required to predict a label and the excess loss introduced by the adaptive system can be recursively defined. As in \cite{wang2015efficient} we can reduce the early exit policy training for minimizing the global risk to a WBC problem. The key idea is that, for each input, a policy must identify whether or not the future reward (expected future accuracy minus comp. loss) outweighs the current-stage accuracy.

To this end, we first focus on the problem of learning the decision function $\gamma_4$, which determines if an example should exit after the fourth convolutional layer or whether it will be classified using the entire network. The time it takes to predict the label of example $x$ depends on this decision and can be written as
\begin{align}\label{eqn.budget_error_tradeoff}
T_4\left(x,\gamma_4\right)=\begin{cases} T+\tau(\gamma_4) & \mbox{if } \gamma_4(\sigma_4(x))=1\\
t_4 + \tau(\gamma_4) & \mbox{otherwise}
\end{cases},
\end{align}
where $\tau(\gamma_4)$ is the computational time required to evaluate the function $\gamma_4$. Our goal is to learn a system that trades-off the evaluation time and the induced error:
\begin{align}
\argmin_{\gamma_4 \in \Gamma}\mathds{E}_{x \sim \mathcal{X}}&\left[T_4(x,\gamma_4)\right] + \lambda \mathds{E}_{(x,y)\sim \mathcal{X}\times\mathcal{Y}}\Big[\bigg(L\left(\hat{y}_4(x),y\right)\notag \\
&-L\left(\hat{y}(x),y\right)\bigg)_{+}\mathds{1}_{\gamma_4(\sigma_4(x))=-1}\Big]
\end{align}
where $(\cdot)_+$ is the function$(z)_+=\max(z,0)$ and $\lambda \in \mathds{R}^+$ is a trade-off parameter that balances between evaluation time and error. Note that the function $T_4\left(x,\gamma_4\right)$ can be expressed as a sum of indicator functions:
\begin{align*}
T_4\left(x,\gamma_4\right)=&\left(T+\tau(\gamma_4)\right)\mathds{1}_{\gamma_4(\sigma_4(x))=1}\\
&\qquad\qquad+\left(t_4+\tau(\gamma_4)\right)\mathds{1}_{\gamma_4(\sigma_4(x))=-1}\\
=&T\mathds{1}_{\gamma_4(\sigma_4(x))=1}+t_4\mathds{1}_{\gamma_4(\sigma_4(x))=-1}+\tau_4(\gamma_4)
\end{align*}

Substituting for $T_4(x,\gamma_4)$ allows us to reduce the problem to an importance weighted binary learning problem:
\begin{align}\label{eqn.single_policy_learn_example}
\argmin_{\gamma_4\in\Gamma}\mathds{E}_{(x,y)\sim\mathcal{X}\times\mathcal{Y}}&\left[C_4(x,y)\mathds{1}_{\gamma_4(\sigma_4(x))\neq \beta_4(x)}\right]+\tau(\gamma_4)
\end{align}
where $\beta_4(x)$ and $C_4(x,y)$ are the optimal decision and cost at stage 4 for the example $(x,y)$ defined:
$$
\beta_4(x)=\begin{cases}-1 & \mbox{if } T>\Big(t_4 + \lambda\Big(L\left(\hat{y}_4(x),y\right)\\
& \qquad\qquad\qquad \qquad -L\left(\hat{y}(x),y\right)\Big)_{+} \Big)\\
1 & \mbox{otherwise}
\end{cases}
$$
and 
$$
C_4(x,y)=\left|T-t_4-\lambda\left(L\left(\hat{y}_4(x),y\right)-L\left(\hat{y}(x),y\right)\right)_{+} \right|.
$$
Note that the regularization term, $\tau(\gamma_4)$, is important to choose the optimal functional form for the function $\gamma_4$ as well as a natural mechanism to define the structure of the early exit system. Rather than limiting the family of function $\Gamma$ to a single functional form such as a linear function or a specific network architecture, we assume the family of functions $\Gamma$ is the union of multiple functional families, notably including the constant decision functions $ \gamma_4(x)=1, \forall x \in |\mathcal{X}|$. Although this constant function does not allow for adaptive network evaluation at the specific location, it additionally does not introduce any computational overhead, that is, $\tau(\gamma_4)=0$. By including this constant function in $\Gamma$, we guarantee that our technique can only decrease the test-time cost.

Empirically, we find that the most effective policies operate on classifier confidences such as classification entropy. Specifically, we consider the family of functions $\Gamma$ as the union of three functional families, the aforementioned constant functions, linear classifier on confidence features generated from linear classifiers applied to $\sigma_4(x)$, and linear classifier on confidence features generated from deep classifiers applied to $\sigma_4(x)$.

Rather than optimizing jointly over all three networks, we leverage the fact that the optimal solution to Eqn. \eqref{eqn.single_policy_learn_example} can be found by optimizing over each of the three families of functions independently. For each family of functions, the policy evaluation time $\tau(\gamma_4)$ is constant, and therefore solving \eqref{eqn.single_policy_learn_example} over a single family of functions is equivalent to solving an unregularized learning problem. We exploit this by solving the three unregularized learning problems and taking the minimum over the three solutions. 

In order to learn the sequence of decision functions, we consider a bottom-up training scheme, as previously proposed in sensor selection \cite{wang2015efficient}. In this scheme, we learn the deepest (in time) early exit block first, then fix the outputs. Fixing the outputs of this trained function, we then train the early exit function immediately preceding the deepest early exit function ($\gamma_3$ in Fig. \ref{fig:network_selection_arch}). 

For a general early exit system, we recursively define the future time, $T_k(x,\gamma_k)$, and the future predicted label, $\tilde{y}_k(x,\gamma_k)$, as
$$
T_{k}(x,\gamma_k)=\\
\begin{cases}
T+\tau(\gamma_k) & \mbox{if } \gamma_k(\sigma_k(x))=1, k=K\\
T_{k+1}(x,\gamma_k& \mbox{if }  \gamma_k(\sigma_k(x))=1, k<K\\
\,\,+1)+\tau(\gamma_k)&\\
t_k + \tau(\gamma_k) & \mbox{otherwise}
\end{cases}
$$
and
$$
\tilde{y}_{k}(x,\gamma_k)=\begin{cases}
\hat{y}(x) & \mbox{if } k=K+1\\
\hat{y}(x) & \mbox{if } k=K \\
& \qquad\mbox{ and } \gamma_k(\sigma_k(x))=1\\
\tilde{y}_{k+1}(x,\gamma_{k+1}) & \mbox{if } k<K \\
& \qquad\mbox{ and } \gamma_k(\sigma_k(x))=-1\\
\hat{y}_{k}(x) & \mbox{otherwise}
\end{cases}.
$$
Using these definitions, we can generalize Eqn. \eqref{eqn.single_policy_learn_example}. For a system with $K$ early exit functions, the $k^{\text{th}}$ early exit function can be trained by solving the supervised learning problem:
\begin{align}\label{eqn.single_policy}
\argmin_{\gamma_k\in\Gamma}\mathds{E}_{(x,y)\sim\mathcal{X}\times\mathcal{Y}}\left[C_k(x,y)\mathds{1}_{\gamma_k(x)\neq \beta_k(\sigma_k(x))}\right]+\tau(\gamma_k),
\end{align}
where optimal decision and cost $\beta_k(x)$ and $C_k(x,y)$ can be defined:
$$
\beta_k(x)=\begin{cases}-1 & \mbox{if } k <K \mbox { and }T_{k+1}(x,\gamma_{k+1})\geq t_k +\\
& \mbox{        } \lambda\left(L\left(\hat{y}_k(x),y\right)-L\left(\tilde{y}_{k+1}(x),y\right)\right)_{+} \\
-1 & \mbox{if } k=K \mbox { and }T\geq t_k + \\
& \mbox{        }\lambda\left(L\left(\hat{y}_k(x),y\right)-L\left(\tilde{y}_{k+1}(x),y\right)\right)_{+}  \\
1 & \mbox{otherwise}
\end{cases}
$$ 
$$
C_k(x,y)=\begin{cases}\Big|T_{k+1}(x,\gamma_{k+1})-t_k & \mbox{if } k<K\\
-\lambda\Big(L\left(\hat{y}_k(x),y\right)-L\left(\tilde{y}_{k+1}(x),y\right)\Big)_{+} \Big|&\\
\Big|T-t_k  & \mbox{otherwise}\\
-\lambda\left(L\left(\hat{y}_k(x),y\right)-L\left(\hat{y}(x),y\right)\right)_{+} \Big|&
\end{cases}.
$$
Eqn. \eqref{eqn.single_policy} allows for efficient training of an early exit system by sequentially training early exit decision functions from the bottom of the network upwards. Furthermore, by including constant functions in the family of functions $\Gamma$ and training early exit functions in all potential stages of the system, the early exit architecture can also naturally be discovered. Finally, in the case of single option at each exit, the layer-wise learning scheme is equivalent to jointly optimizing all the exits with respect to full system risk. 

\section{Network Selection}

As shown in Fig. \ref{fig:imagenet_challenge}, the computational time has grown dramatically with respect to classification performance. Rather than attempting to reduce the complexity of the state-of-the-art networks, we instead leverage this non-linear growth by extending the early exiting strategy to the regime of network selection. Conceptually, we seek to exploit the fact that many examples are correctly classified by relatively efficient networks such as alexnet and googlenet, whereas only a small fraction of examples are correctly classified by computationally expensive networks such as resnet 152 and incorrectly classified by googlenet and alexnet.

As an example, assume we have three pre-trained networks, $N_1$, $N_2$, and $N_3$. For an example $x$, denote the predictions for the networks as $N_1(x)$, $N_2(x)$, and $N_3(x)$. Additionally, denote the evaluation times for each of the networks as $\tau(N_1)$, $\tau(N_2)$, and $\tau(N_3)$. 

As in Fig. \ref{fig:network_selection_arch}, the adaptive system composed of two decision functions that determine which network is evaluated for each example. First, $\kappa_1:|\mathcal{X}|\rightarrow \{N_1,N_2,N_3\}$ is applied after evaluation of $N_1$ to determine if the classification decision from $N_1$ should be returned or if network $N_2$ or network $N_3$ should be evaluated for the example. For examples that are evaluated on $N_2$, $\kappa_2:|\mathcal{X}|\rightarrow \{N_2,N_3\}$ determines if the classification decision from $N_2$ should be returned or if network $N_3$ should be evaluated.%

Our goal is to learn the functions $\kappa_1$ and $\kappa_2$ that minimize the average evaluation time subject to a constraint on the average loss induced by adaptive network selection. As in the adaptive early exit case, we first learn $\kappa_2$ to trade-off between the average evaluation time and induced error:
\begin{align}
&\min_{\kappa_2 \in \Gamma}\mathds{E}_{x \sim \mathcal{X}}\left[\tau(N_3)\mathds{1}_{\kappa_2(x)=N_3}\right]+\tau(\kappa_2)\notag\\
&\qquad+\lambda\mathds{E}_{(x,y)\sim \mathcal{X}\times\mathcal{Y}}\Bigg[\Big(L\left(N_2(x),y\right)\notag\\
&\qquad \qquad-L\left(N_3(x),y\right)\Big)_{+}\mathds{1}_{\kappa_2(x)=N_2}\Bigg],
\end{align}
where $\lambda\in \mathds{R}^+$ is a trade-off parameter. As in the adaptive network usage case, this problem can be posed as an importance weighted supervised learning problem:
\begin{align}\label{eqn.kappa2_training}
\min_{\kappa_2\in\Gamma}\mathds{E}_{(x,y)\sim\mathcal{X}\times\mathcal{Y}}\left[W_2(x,y)\mathds{1}_{\kappa_2(x)\neq \theta_2(x)}\right]+\tau(\kappa_2),
\end{align}
where $\theta_2(x)$ and $W_2(x,y)$ are the cost and optimal decision at stage 4 for the example/label pair $(x,y)$ defined:
$$
\theta_2(x)=\begin{cases}N_2& \mbox{if}\, \tau(N_3)>\lambda\left(L\left(N_3(x),y\right)-L\left(N_2(x),y\right)\right)_{+} \\
N_3 & \mbox{otherwise}
\end{cases}
$$
and 
\begin{align*}
W_2(x,y)=\Big|&\tau(N_3)-\lambda\left(L\left(N_2(x),y\right)-L\left(N_3(x),y\right)\right)_{+} \Big|.
\end{align*}
Once $\kappa_2$ has been trained according to Eqn. \eqref{eqn.kappa2_training}, the training times for examples that pass through $N_2$ and are routed by $\kappa_2$ can be defined $T_{\kappa_2}(x)=\tau(N_2)+\tau(\kappa_2)+\tau(N_3)\mathds{1}_{\kappa_2(x)=N_3}$. As in the adaptive early exit case, we train and fix the last decision function, $\kappa_2$, then train the earlier function, $\kappa_1$. As before, we seek to trade-off between evaluation time and error:
\begin{align}
&\min_{\kappa_1 \in \Gamma}\mathds{E}_{x \sim \mathcal{X}}\left[\tau(N_3)\mathds{1}_{\kappa_1(x)=N_3}+\tau(N_2)\mathds{1}_{\kappa_1=N_2}\right]+\tau(\kappa_1)+\notag\\
&\lambda\mathds{E}_{(x,y)\sim \mathcal{X}\times\mathcal{Y}}\Bigg[\left(L\left(N_2(x),y\right)-L\left(N_3(x),y\right)\right)_{+}\mathds{1}_{\kappa_1(x)=N_2}\notag\\
&+\left(L\left(N_1(x),y\right)-L\left(N_3(x),y\right)\right)_{+}\mathds{1}_{\kappa_1(x)=N_1}\Bigg]
\end{align}
This can be reduced to a cost sensitive learning problem:
\begin{align}\label{eqn.kappa1_training}
\min_{\kappa_1\in\Gamma}\mathds{E}_{(x,y)\sim\mathcal{X}\times\mathcal{Y}}\Bigg[&R_3(x,y)\mathds{1}_{\kappa_1(x)=N_3}+R_2(x,y)\mathds{1}_{\kappa_1(x)=N_2}\notag\\
+&R_1(x,y)\mathds{1}_{\kappa_1(x)=N_1}\Bigg]+\tau(\kappa_1),
\end{align}
where the costs are defined:
\begin{align*}
R_1(x,y)&=\left(L(N_1(x),y)-L(N_3(x),y)\right)_+\notag\\
R_2(x,y)&=\left(L(N_2(x),y)-L(N_3(x),y)\right)_+ + \tau(N_2)\notag\\
R_3(x,y)&=\tau(N_3)\notag.
\end{align*}
\setlength{\textfloatsep}{5pt}
\begin{algorithm}[H]
    \begin{algorithmic}
   \STATE {\bfseries Input:} Data: $(x_i,y_i)_{i=1}^n$,
   \STATE Models $\mathcal{S}$, routes, $E$, model costs $\tau(.)$), 
   \WHILE{ $\exists$ untrained $\pi$}
   \STATE ({\bf1}) Choose the deepest policy decision $j$, s.t. all down-stream policies are trained
   \FOR{example $i \in \{1,\ldots,n\}$}
   \STATE ({\bf2}) Construct the weight vector $\vec{w}_i$ of costs per action from Eqn. \eqref{eqn.kappa2_training}.
   \ENDFOR
   \STATE ({\bf3}) $\pi_j \leftarrow$Learn clf.$\left((x_1,\vec{w}_1),\ldots,(x_n,\vec{w}_n) \right )$
   \STATE ({\bf4}) Evaluate $\pi_j$ and update route costs to model $j$:
   \STATE $C(x_i,y_i,s_n,s_j) \leftarrow \vec{w}^j_i\left(\pi_j(x_i)\right)+C(x_i,y_i,s_n,s_j)$
   \ENDWHILE
   \STATE ({\bf5}) Prune models the policy does not route any example to from the collection
   \STATE {\bfseries Output:} Policy functions, $\pi_1,\ldots,\pi_K$
    \end{algorithmic}
    \caption{Adaptive Network Learning Pseudocode}
    \label{ns_algorithm}
\end{algorithm}
\section{Experimental Section}

\begin{figure*}[!t]
	\vspace{-8pt}
   \begin{center}
    \includegraphics[width=0.49\linewidth]{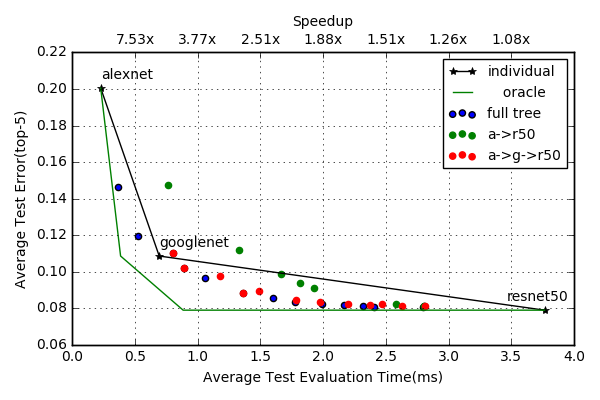}
    \includegraphics[width=0.50\linewidth]{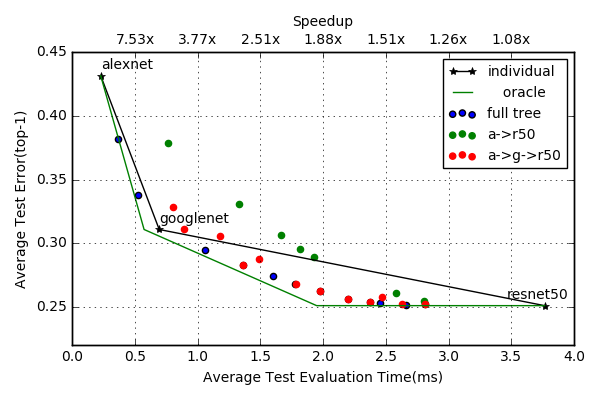}
  \end{center}
  \vspace{-15pt}
  \caption[Network selection performance]{\small Performance of network selection policy on Imagenet (Left: top-5 error Right: top-1 error). Our full adaptive system (denoted with blue dots) significantly outperforms any individual network for almost all budget regions and is close to the performance of the oracle. The performances are reported on the validation set of ImageNet dataset.}
  \label{fig:network_selection_perf_top5}
\end{figure*}
\begin{figure*}[!t]
   \vspace{-10pt}
   \begin{center}
    \includegraphics[width=0.28\linewidth]{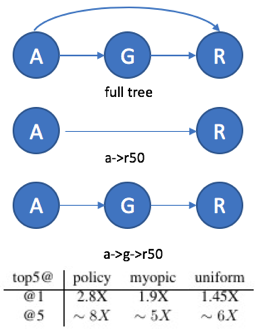}
    \includegraphics[width=0.58\linewidth]{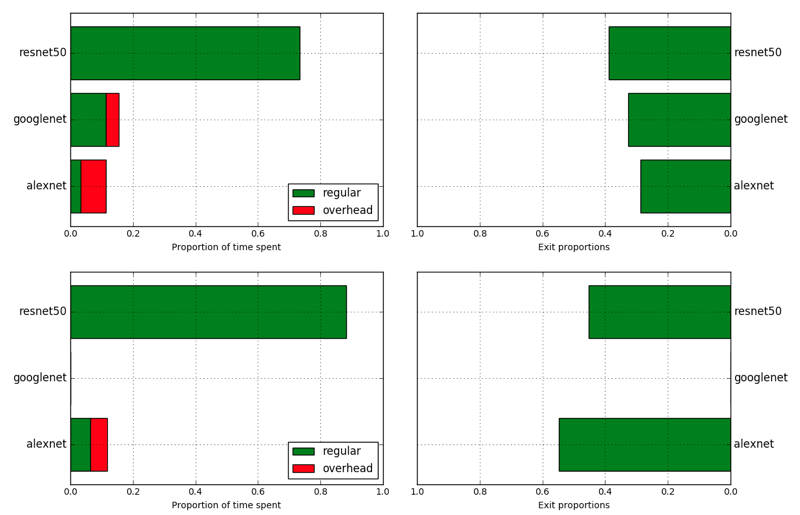}
  \end{center}
  \vspace{-10pt}
  \caption[Network usage analysis]{\small \textbf{(Left)} Different network selection topologies that we considered. Arrows denote possible jumps allowed to the policy. A, G and R denote Alexnet, GoogLeNet and Resnet50, respectively. \textbf{(Right)} Statistics for proportion of total time spent on different networks and proportion of samples that exit at each network. Top row is sampled at 2.0ms and bottom row is sampled at 2.8ms system evaluation}
  \vspace{-5pt}
  \label{fig:network_usage_analysis}
\end{figure*}

We evaluate our method on the Imagenet 2012 classification dataset \cite{russakovsky2015imagenet} which has 1000 object classes. We train using the 1.28 million training images and evaluate the system using 50k validation images. We use the pre-trained models from Caffe Model Zoo for Alexnet, GoogLeNet and Resnet50 \cite{krizhevsky2012imagenet, szegedy2015going,he2016deep}. For preprocessing we follow the same routines proposed for these networks and verify the final network performances within a small margin ($<0.1\%$). Note that it is common to use ensembles of networks and multiple crops to achieve maximum performance. These methods add minimal gain in accuracy while increasing the system cost dramatically. As the speedup margin increases, it becomes trivial for the policy to show significant speedups within the same accuracy tolerance. We believe such speedups are not useful in practice and focus on single crop with single model case.

\emph{Temporal measurements:}
We measure network times using the built-in tool in the Caffe library on a server that utilizes a Nvidia Titan X Pascal with CuDNN 5. Since our focus is on the computational cost of the networks, we ignore the data loading and preprocessing times. The reported times are actual measurements including the policy overhead.

\emph{Policy form and meta-features:}
In addition to the outputs of the convolutional layers of earlier networks, we augment the feature space with the entropy of prediction probabilities. We relax the indicators in equations \eqref{eqn.single_policy} and \eqref{eqn.kappa1_training} learn  linear logistic regression model on these features for our policy.
We experimented with pooled internal representations, but in practice, inclusion of the entropy feature with a simple linear policy significantly outperforms more complex policy functions that exclude the entropy feature. %
\subsection{Network Selection}

\emph{Baselines:}
Our full system, depicted in Figure \ref{fig:network_selection_arch}, starts with Alexnet. Following the evaluation of Alexnet, the system determines for each example either to return the prediction, route the example to GoogLeNet, or route the example to Resnet50. For examples that are routed to GoogLeNet, the system either returns the prediction output by GoogLeNet or routes the example to Resnet50. As baselines, we compare against a uniform policy and a myopic policy which learns a single threshold based on model confidence. We also report performance from different system topologies. To provide a bound on the achievable performance, we show the performance of a soft oracle. The soft oracle has access to classification labels and sends each example to the fastest model that correctly classifies the example. Since having access to the labels is too strong, we made the oracle softer by adding two constraints. First, it follows the same network topology, also it can not make decisions without observing the model feedback first, getting hit by the same overhead. Second, it can only exit at a cheaper model if all latter models agree on the true label. This second constraint is added due to the fact that our goal is not to improve the prediction performance of the system but to reduce the computational time, and therefore we prevent the oracle from ``correcting'' mistakes made by the most complex networks.

We sweep the cost trade-off parameter in the range 0.0 to 0.1 to achieve different budget points. Note that due to weights in our cost formulation, even when the pseudo labels are identical, policy behavior can differ. Conceptually, the weights balance the importance of the samples that gain in classification loss in future stages versus samples that gain in computational savings by exiting early stages.

The results are demonstrated in Figure \ref{fig:network_selection_perf_top5}. We see that both full tree and a->g->r50 cascade achieve significant (2.8x) speedup over using Resnet50 while maintaining its accuracy within $1\%$. The classifier feedback for the policy has a dramatic impact on its performance. Although, Alexnet introduces much less overhead compared to GoogLeNet ($\approx$0.2 vs $\approx$0.7), the a->r50 policy performs significantly worse in lower budget regions. Our full tree policy learns to choose the best order for all budget regions. Furthermore, the policy matches the soft oracle performance in both the high and low budget regions.

Note that GoogLeNet is a very well positioned at 0.7ms per image budget, probably due to its efficiency oriented architectural design with inception blocks \cite{szegedy2015going}. For low budget regions, the overhead of the policy is a detriment, as even when it can learn to send almost half the samples to Alexnet instead of GoogLeNet with marginal loss in accuracy, the extra 0.23ms Alexnet overhead brings the balance point, $\approx 0.65ms$, very close to using only GoogLeNet at 0.7ms. The ratio between network evaluation times is a significant factor for our system. Fortunately, as mentioned before, for many applications the ratio of different models can be very high (cloud computing upload times, resnet versus Alexnet difference etc.).

We further analyzed the network usage and runtime proportion statistics for samples at different budget regions. Fig. \ref{fig:network_usage_analysis} demonstrates the results at three different budget levels. Full tree policy avoids using GoogLeNet altogether for high budget regions. This is the expected behavior since the a->r50 policy performs just as well in those regions and using GoogLeNet in the decision adds too much overhead. 
At mid level budgets the policy distributes samples more evenly. Note that the sum of the overheads is close to useful runtime of cheaper networks in this region. This is possible  since the earlier networks are very lightweight.

\subsection{Network Early Exits}

To output a prediction following each convolutional layer, we train a single layer linear classifier after a global average pooling for each layer. We added global pooling to minimize the policy overhead in earlier exits.
For Resnet50 we added an exit after output layers of 2a, 2c, 3a, 3d, 4a and 4f. The dimensionality of the exit features after global average pooling are 256, 256, 512, 512, 1024 and 1024 in the same order as the layer names.
For GoogLeNet we added the exits after concatenated outputs of every inception layer.

\begin{figure}[h]
	\vspace{-13pt}
   \begin{center}
    \includegraphics[width=\linewidth]{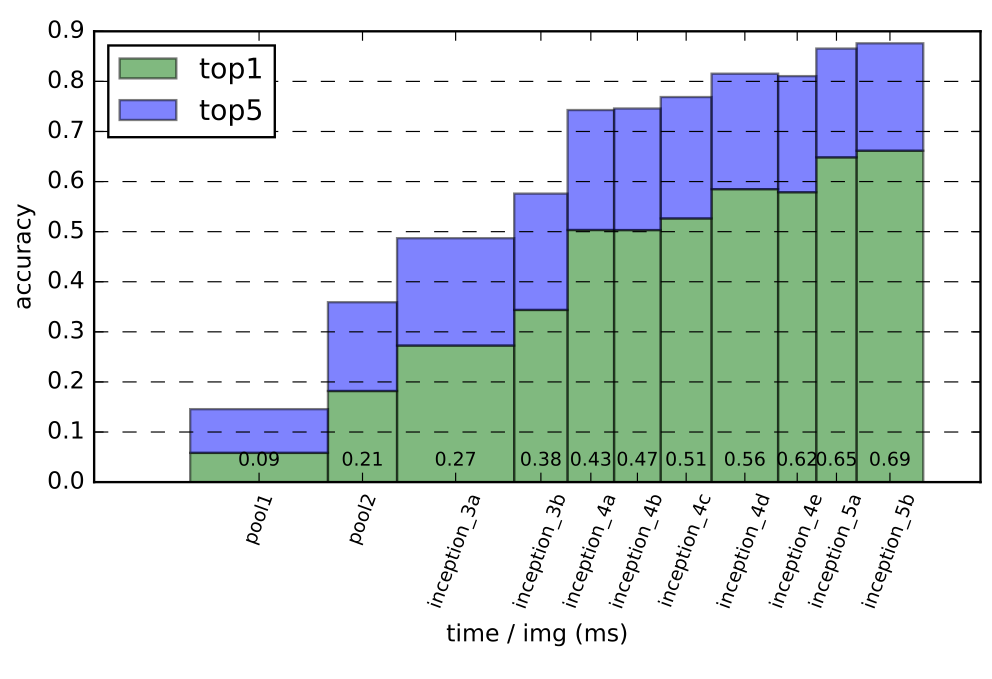}
    \includegraphics[width=\linewidth]{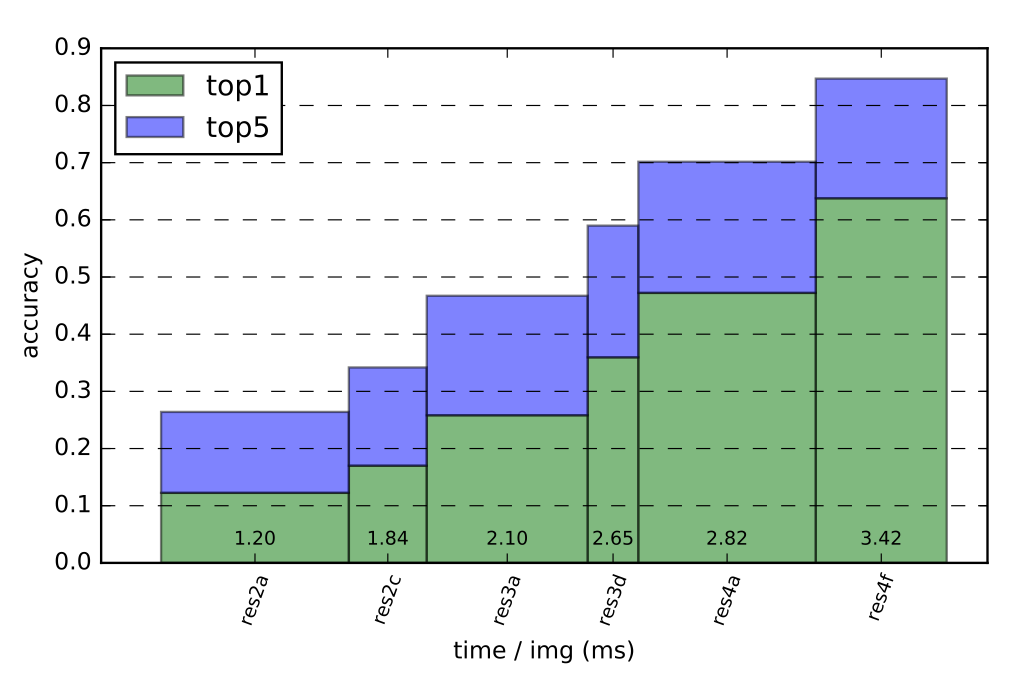}
  \end{center}
  \vspace{-20pt}
  \caption[Early exit analysis]{ \small The plots show the accuracy gains at different layers for early exits for networks GoogLeNet (top) and Resnet50 (bottom).}
  \label{fig:early_exit_analysis}
\end{figure}

Table \ref{table:early_exit_results} shows the early exit performance for different networks. The gains are more marginal compared to network selection.
Fig \ref{fig:early_exit_analysis} shows the accuracy gains per evaluation time for different layers.
Interestingly, the accuracy gain per time is more linear within the same architecture compared to different network architectures. This explains why the adaptive policy works better for network selection compared to early exits.

\begin{table}
\begin{center}
\begin{tabular}{ l | c | c c}
  Network & policy top-5 & uniform top-5 & \\
  \hline
  GoogLeNet@1 & $9\%$ & $2\%$ & \\
  GoogLeNet@2 & $22\%$ & $9\%$ &\\
  GoogLeNet@5 & $33\%$ & $20\%$ &  \\
  \hline
  Resnet50@1 & $8\%$ & $1\%$  & \\
  Resnet50@2 & $18\%$ & $12\%$  & \\
  Resnet50@5 & $22\%$ & $10\%$ & \\
\end{tabular}
\end{center}
\vspace{-10pt}
\caption{Early exit performances at different accuracy/budget trade-offs for different networks. @x denotes x loss from full model accuracy and reported numbers are percentage speed-ups. \label{table:early_exit_results}}
\vspace{-0.1cm}
\end{table}

\begin{figure}[!t]
	\vspace{-5pt}
    \includegraphics[width=0.495\linewidth]{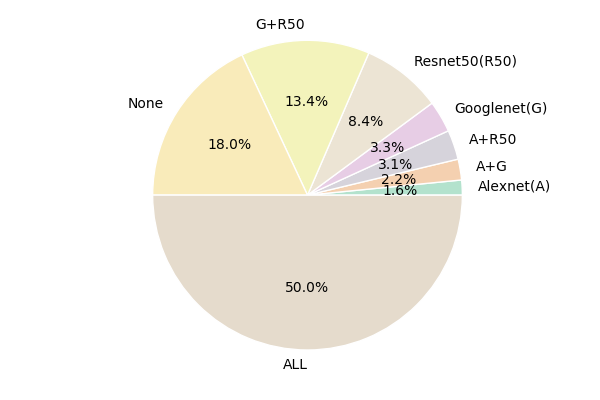}
    \includegraphics[width=0.495\linewidth]{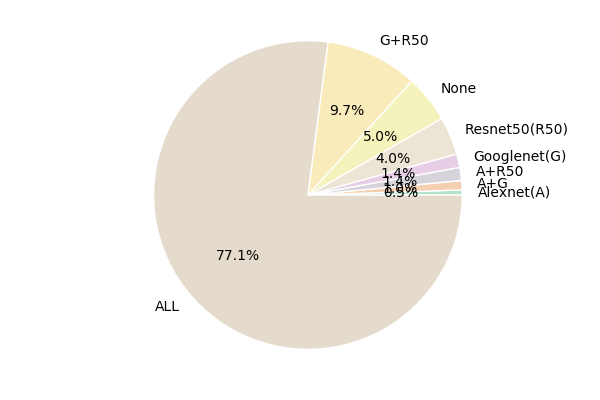}
   \vspace{-0.8cm}
  \caption{\small Analysis of top-1 and top-5 errors for different networks. Majority of the samples are easily classified by Alexnet, and only a minority of them require deeper networks.}
  \label{fig:error_analysis_top1}
  \vspace{-0.1cm}
\end{figure}

\subsection{Network Error Analysis}
Fig. \ref{fig:error_analysis_top1} shows the distributions over examples of the networks that correctly label the example. Notably, $50\%$ and $77\%$ of the examples are correctly classified by all networks for top 1 and top 5 error, respectively. Similarly, $18\%$ and $5\%$ of the examples are incorrectly classified by all networks with respect to their top 1 and top 5 error, respectively. These results verify our hypothesis that for a large fraction of data, there is no need for costly networks. In particular, for the $68\%$ and $82\%$ of data with no change in top 1 and top 5 error, respectively, the use of any network apart from Alexnet is unnecessary and only adds unnecessary computational time.

Additionally, it is worth noting the balance between examples incorrectly classified by all networks, $18\%$ and $5\%$ respectively for top 1 and top 5 error, and the fraction of examples correctly classified by either GoogLeNet or Resnet but not Alexnet, $25.1\%$ and $15.1\%$ for top 1 and top 5 error, respectively. This behavior supports our observation that entropy of classification decisions is an important feature in making policy decisions, as examples likely to be incorrectly classified by Alexnet are likely to be classified correctly by a later network.

Note that our system is trained using the same data used to train the networks. Generally, the resulting evaluation error for each network on training data is significantly lower than error that arises on test data, and therefore our system is biased towards sending examples to more complex networks that generally show negligible training error. Practically, this problem is alleviated through the use of validation data to train the adaptive systems. In order to maintain the reported performance of the network without expansion of the training set, we instead utilize the same data for training both networks and adaptive systems, however we note that performance of our adaptive systems is generally better when trained on data excluded from the network training.

\section{Conclusion}
We proposed two different schemes to adaptively trade off model accuracy with model evaluation time for deep neural networks. We demonstrated that significant gains in computational time is possible through our novel policy with negligible loss in accuracy on ImageNet image recognition dataset. We posed a global objective for learning an adaptive early exit or network selection policy and solved it by reducing the policy learning problem to a layer-by-layer weighted binary classification problem. We believe that adaptivity is very important in the age of growing data for models with high variance in computational time and quality. We also showed that our method approximates an Oracle based policy that has benefit of access to true error for each instance from all the networks. 

\section*{Acknowledgements} 
This material is based upon work supported in part by NSF Grants CCF: 1320566, NSF Grant CNS: 1330008 NSF CCF: 1527618, the U.S. Department of Homeland Security, Science and Technology Directorate, Office of University Programs, under Grant Award 2013-ST-061-ED0001, and by ONR contract N00014-13-C-0288. The views and conclusions contained in this document are those of the authors and should not be interpreted as necessarily representing the social policies, either expressed or implied, of the NSF, U.S. DHS, ONR or AF. 


\bibliography{ann}
\bibliographystyle{icml2017}

\end{document}